\documentclass{article}
\usepackage[utf8]{inputenc}
\usepackage{authblk}
\usepackage{setspace}
\usepackage[margin=1.25in]{geometry}
\usepackage{graphicx}
\graphicspath{ {./figures/} }
\usepackage{subcaption}
\usepackage{amsmath}
\usepackage{lineno}
\usepackage{verbatim}
\usepackage{makecell}
\usepackage{float}
\usepackage{booktabs}
\usepackage{changepage}

\usepackage{cleveref}
\usepackage{natbib}
\title{Assesment of material layers in building walls using GeoRadar}

\author[1]{Ildar Gilmutdinov$^{1}$}
\author[2]{Ingrid Schlögel$^{2}$}
\author[3]{Alois Hinterleitner$^{2}$}
\author[4]{Peter Wonka$^{3}$}
\author[5]{Michael Wimmer$^{1}$}

\affil[1]{TU Wien}
\affil[2]{ZAMG (Zentralanstalt für Meteorologie und Geodynamik)}
\affil[3]{KAUST (King Abdullah University of Science and Technology)}

\date{}

\begin{document}

\maketitle

\begin{abstract}
Assessing the structure of a building with non-invasive methods is an important problem. One of the possible approaches is to use GeoRadar to examine wall structures by analyzing the data obtained from the scans. We propose a data-driven approach to evaluate the material composition of a wall from its GPR radargrams. In order to generate training data, we use gprMax to model the scanning process. Using simulation data, we use a convolutional neural network to predict the thicknesses and dielectric properties of walls per layer. We evaluate the generalization abilities of the trained model on data collected from real buildings.
\end{abstract}

\section{Introduction}

Assessment of existing buildings' recycling costs often requires a destructive method to look through its building elements like walls and floors. In order to answer what materials comprise a wall, one often has to obtain an explicit overview of the cross-section - either by drilling or carving out a piece. Ground-penetrating radar (GPR) presents the way to examine the walls without a destructive invasive process. GeoRadar has been successfully utilized in other fields, such as archaeology \cite{ZHAO2013107}, seismology \cite{zheng} and civil engineering for non-destructive examination \cite{morris}.

In order to identify materials in layered structures, one could refer to the research for a similar problem. Namely, obtaining permittivity maps from the GPR radargrams \cite{pinet}. Previous research concentrated on using machine learning to invert radargrams into permittivity maps, i.e. to receive per-pixel permittivity values as a reconstruction of the original scene. \cite{9321540} used Deep Neural Networks to reconstruct tunnel linings, while \cite{pinet} showed improvement of the idea and its application to more general sub-surface structures. 

\cite{li} used deep learning to estimate the size parameters of reinforcement bars by combining GPR and electromagnetic induction. \cite{liu} used convolutional neural networks (CNN) to invert electrical resistivity data. \cite{zheng} used CNNs for interpretation of seismic images. Aside from deep learning methods, there is another representative of the family of data-driven approaches called Full Wave Inversion (FWI). By minimizing the differences between recorded and modeled waveforms, it adjusts the model to fit the observations \cite{kruk}. 

As an alternative to a grid representation of the permittivity map, we propose to represent it as a sequence of layers. We predict the width and the permittivity of the underlying material for each layer by using a convolutional neural network (CNN). The proposed CNN was trained and evaluated on a dataset of simulated B-scans. Moreover, we collected radargrams of the real buildings and validated the prosed CNN on this dataset.


\section{Methodology}

\begin{figure}[H]
\captionsetup{justification=centering}
\centering
\includegraphics[width=0.7\textwidth]{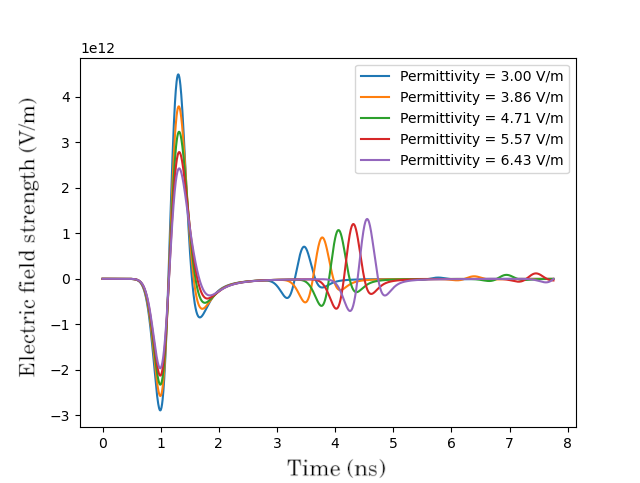}
\caption{Simulated A-scans of a material block with different relative permittivities. Higher relative permittivity increases the wave's travel time.}
\label{fig:perm_change}
\end{figure}

\begin{figure}[H]
\captionsetup{justification=centering}
\centering
\includegraphics[width=0.7\textwidth]{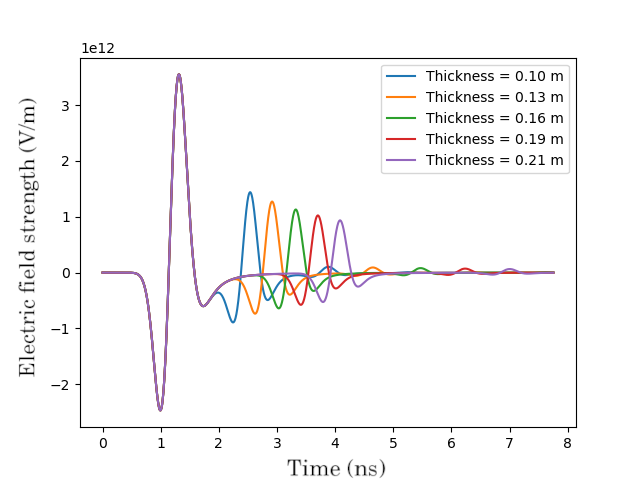}
\caption{Simulated A-scans of a material block with different thicknesses. Thicker blocks imply longer traveling times. }
\label{fig:width_change}
\end{figure}

\begin{figure}[H]
\captionsetup{justification=centering}
\centering
\includegraphics[width=0.7\textwidth]{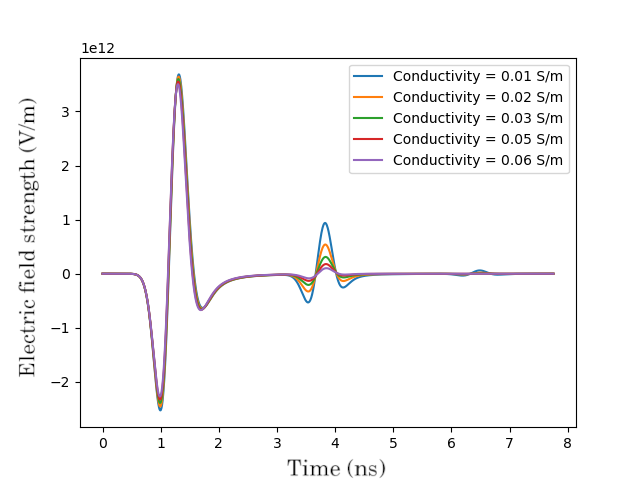}
\caption{Simulated A-scans of a material block with different conductivities. Higher conductivity mostly affects the amplitudes of the peaks and not the wave's velocity.}
\label{fig:cond_change}
\end{figure}

The manual process of analyzing velocities is laborious. By estimating the traveling velocities of the waves, one can estimate the dielectric properties of the material in the examined object. Objects of different permittivities and sizes will affect the traveling velocity of the source wave. Such behavior can be seen in figures \ref{fig:perm_change}-\ref{fig:cond_change}, that show the results of scanning simulations in a setup with a solid material block. The figures \ref{fig:perm_change} and \ref{fig:width_change} demonstrate the effect of a permittivity change and a thickness change, which can be observed to strongly influence the wave's velocities. The change in conductivity mostly affects the amplitudes and not the velocities as it can be seein in Figure \ref{fig:cond_change}.

In this work, we employ a Convolutional Neural Network (CNN) as a model for predicting material layers of walls from the B-scans. Namely, thicknesses and relative permittivities of each layer. CNNs have been performing very successfully in tasks of computer vision and image classification \cite{FANG2020100980,sahiner}. Neural networks approximate a function by constructing a sequence of parameterized computations. Its parameters or weights are adjusted through an iterative process that minimizes the goal function that encodes an objective. 

\subsection{CNN architecture}\label{cnn}

\begin{figure}[h]
    \centering
    \includegraphics[width=0.9\textwidth]{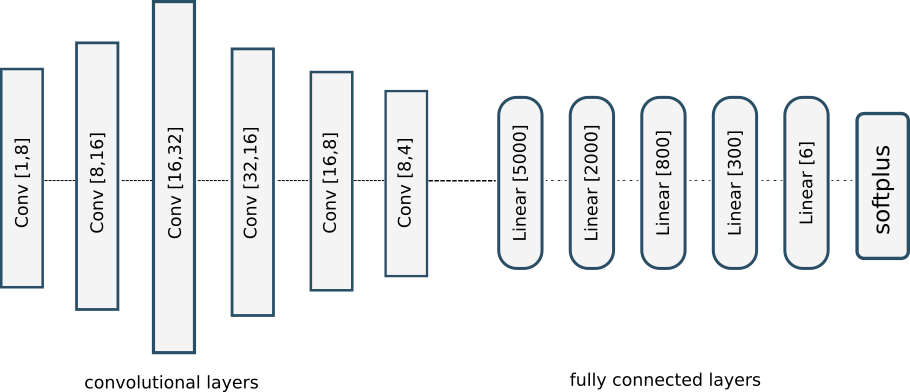}
    \caption{Schematic description of the employed CNN model. Output of six convolutional layers is flattened and passed as input to a series of five fully connected linear layers. \textit{Softplus} enforces positive values on the output.}
    \label{fig:cnn}
\end{figure}

We employ a convolutional neural network (CNN) that takes as input B-scans. Here, B-scans are represented by 2-dimensional arrays of size 255x40. They are then passed to a sequence of convolutional maps of a kernel size of 20x5. We employ six blocks of convolutional layers that are activated with ReLU function. ReLU (Rectified Linear Unit) has been recognized as one of the best choices for the activation function \cite{pmlr-v15-glorot11a}. It is defined as the positive truncate of the incoming argument: $\mathbf{ReLU(x) = max(0,x)}$. The result is then batch normalized in order to stabilize the network's training by re-centering and re-scaling the output of the previous layer. The number of the output feature maps in each such block differs slightly. We use the following sizes: 8, 16, 32, 16, 8, 4, like depicted in Fig. \ref{fig:cnn}. 

The output of the last convolutional layer is then flattened and processed by five fully connected linear layers activated with hyperbolic tangents. Hyperbolic tangent is another common activation function, which showed better performance in our experiments. The sizes of the linear layers were set as 5000, 2000, 800, 300, 12. The final layer has two groups of neurons. The first 6 neurons in the final layer encode thicknesses of the material layers, whereas the next 6 neurons are responsible for the permittivities. In order to ensure positive output values, we employ a softplus function, which is a smooth approximation of the ReLU function. The difference between outputs of the model and corresponding targets from the datasets are computed with the L1 loss that computes an absolute norm of the difference. That is then used to compute the gradients and optimize network parameters. 

\section{Simulated dataset 2D}

Our simulated dataset contains 30 000 simulated GPR B-Scans. For the simulation we used GprMax software \cite{gprmax}. It implements a finite difference time domain method. We set up a 2D grid that is 24cm wide and 46cm thick. It is filled with solid material blocks, whose total thickness is up to 46cm. The set of used relative permittivities is 1 to 7, which represents a typical range for the construction materials. Another dielectric parameter - conductivity, is not considered in this work, since it mostly affects amplitudes which can be visible in Fig. \ref{fig:cond_change}, whereas the important characteristics of a material are determined by the traveling time of the transmitted wave through that medium.

The transmitter and receiver antennas were set 40mm apart, with the transmitting frequency being 1 GHz and the source waveform being the Ricker wavelet \cite{ricker}. The transmitting time window is 12 nanoseconds at a fixed position. After the simulation is complete, we extract the electrical component in the transverse direction of the simulated electromagnetic field and save it as a B-scan. Together with the information about the material class and thickness of the material block, it forms a sample of the dataset.

\begin{figure}[h]
    \centering
    \includegraphics[width=0.4\textwidth]{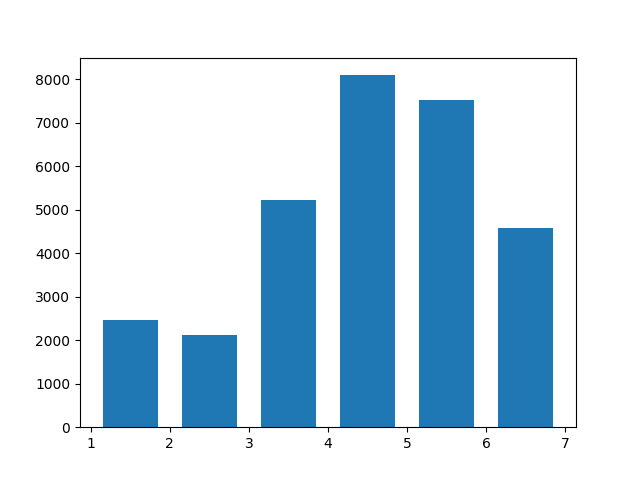}
    \caption{Distribution of the number of layers in samples within the simulated dataset. We favor cases with more layers as they are more difficult to predict. }
    \label{fig:num_of_layers_distr}
\end{figure}

The corresponding targets in the dataset - the walls' configurations - are represented by a stack of up to 6 rectangular blocks. The combinations of different material layers are generated randomly. First, we uniformly sample the number of layers. Next, we sample thicknesses and permittivities from a multinomial distribution, which is a generalization of a binomial distribution. It allows us to sample several variables (thicknesses) at once and avoid the issue when most of the given volume is taken by the first layers. Since there are up to 6 layers, there are 12 variables encoding a target in a configuration: 6 for thickness and 6 for relative permittivity. Also, since the prediction of the further layers is more difficult, we filled the dataset with more instances of the more layered structures, as can be seen from the distribution of the samples over the number of layers in Fig. \ref{fig:num_of_layers_distr}.

In order to consider conditions closer to an actual environment, we add noise grains that vary in size and permittivity. The two small dots at the top are transmitter and receiver antennas. The blocks of shades of blue are material layers, whereas the small colored objects are the noise grains.

\begin{figure}[h!]
    \centering
    \begin{subfigure}{0.3\textwidth}
    \captionsetup{justification=centering}
    \includegraphics[width=\textwidth]{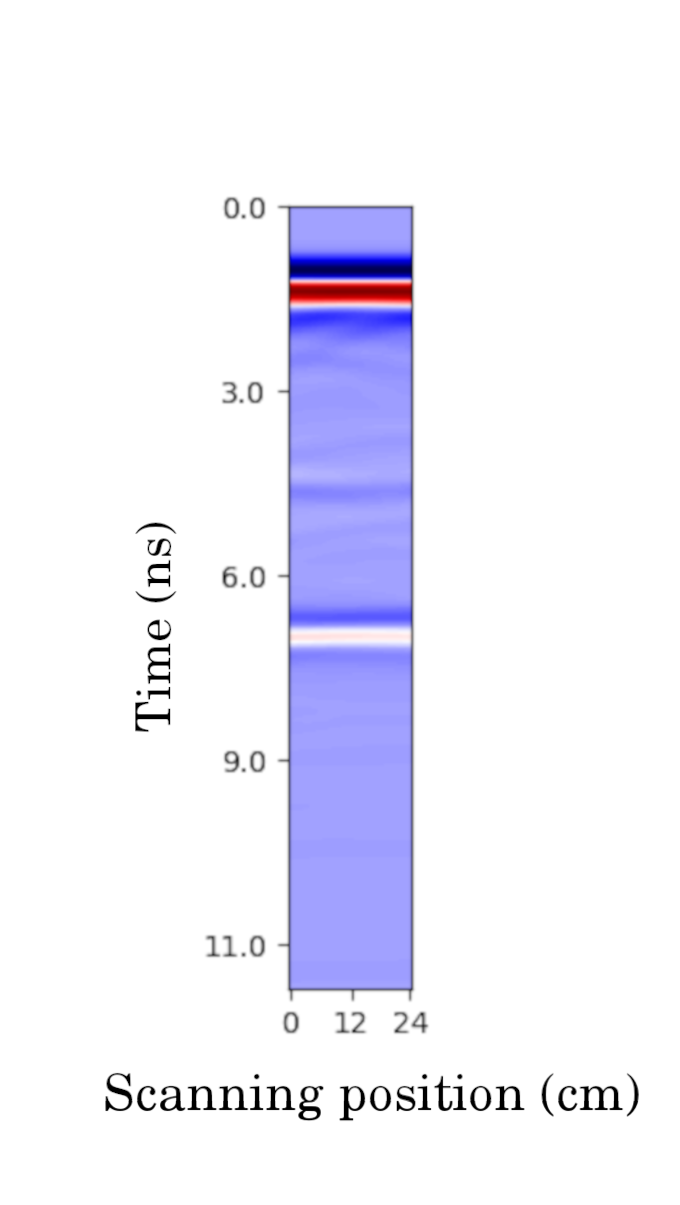}
    \caption{B-Scan 1}
    \label{fig:b_scan_1}
    \end{subfigure}
    \begin{subfigure}{0.3\textwidth}
    \captionsetup{justification=centering}
    \includegraphics[width=\textwidth]{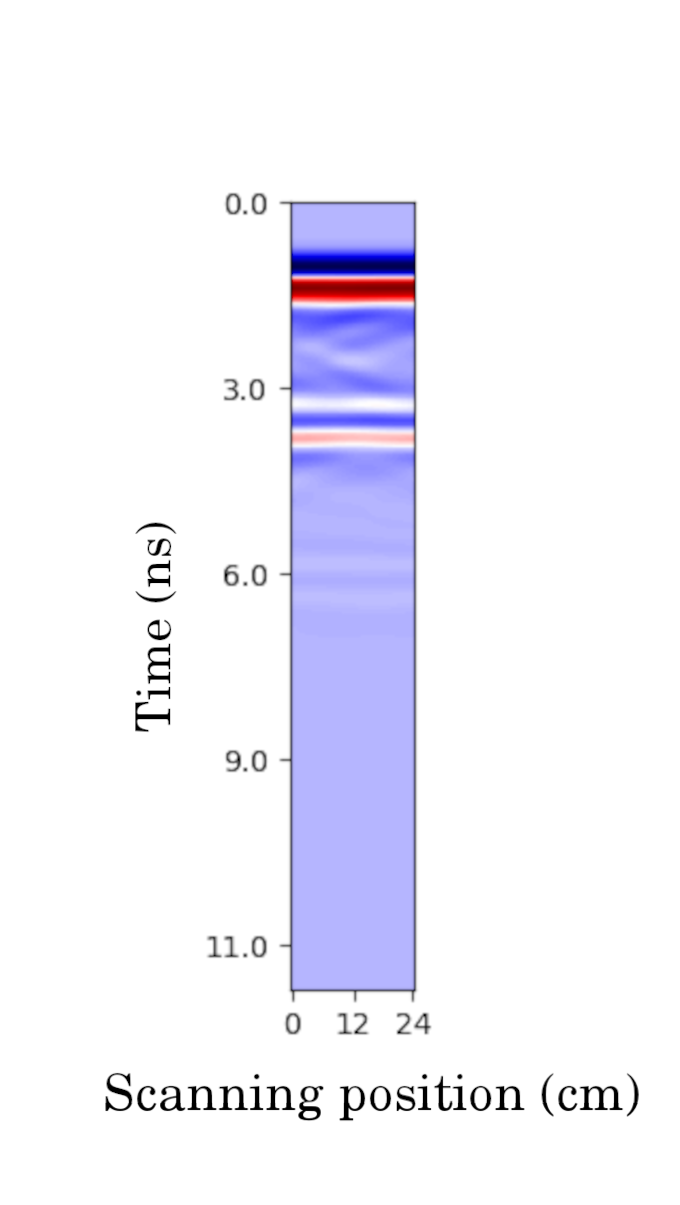}
    \caption{B-Scan 2}
    \label{fig:b_scan_2}
    \end{subfigure}
    \begin{subfigure}{0.3\textwidth}
    \captionsetup{justification=centering}
    \includegraphics[width=\textwidth]{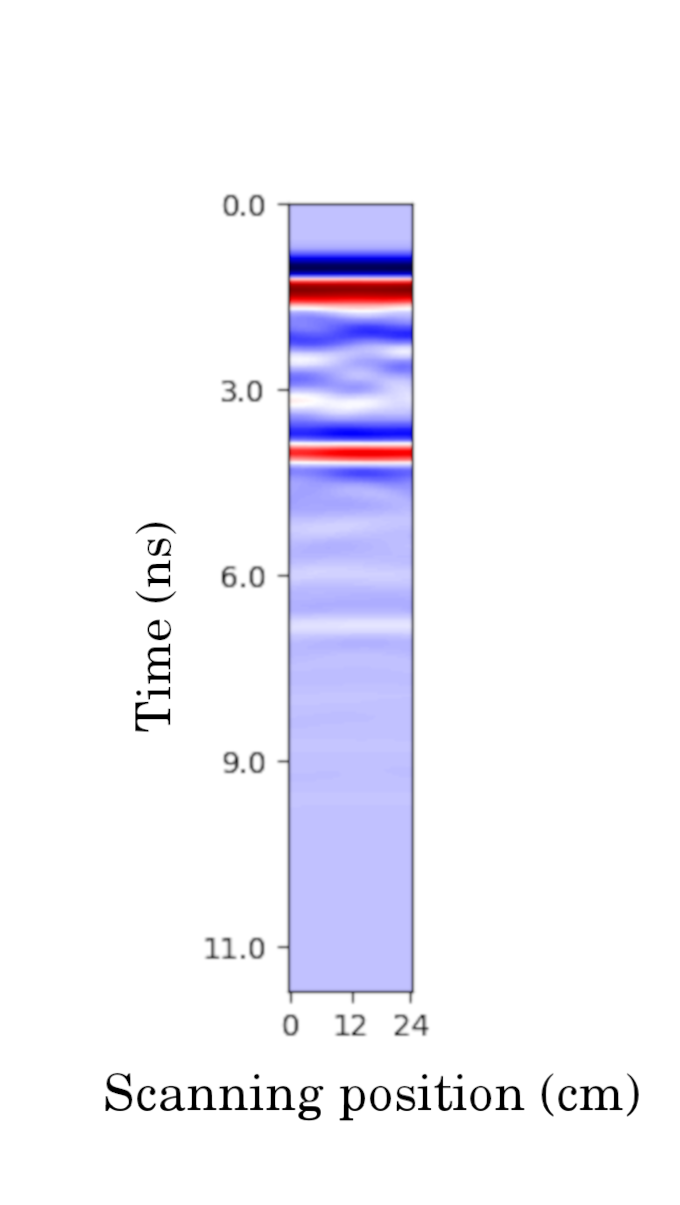}
    \caption{B-Scan 3}
    \label{fig:b_scan_3}
    \end{subfigure}
    \vskip\baselineskip
    \begin{subfigure}{0.3\textwidth}
    \captionsetup{justification=centering}
    \includegraphics[width=\textwidth]{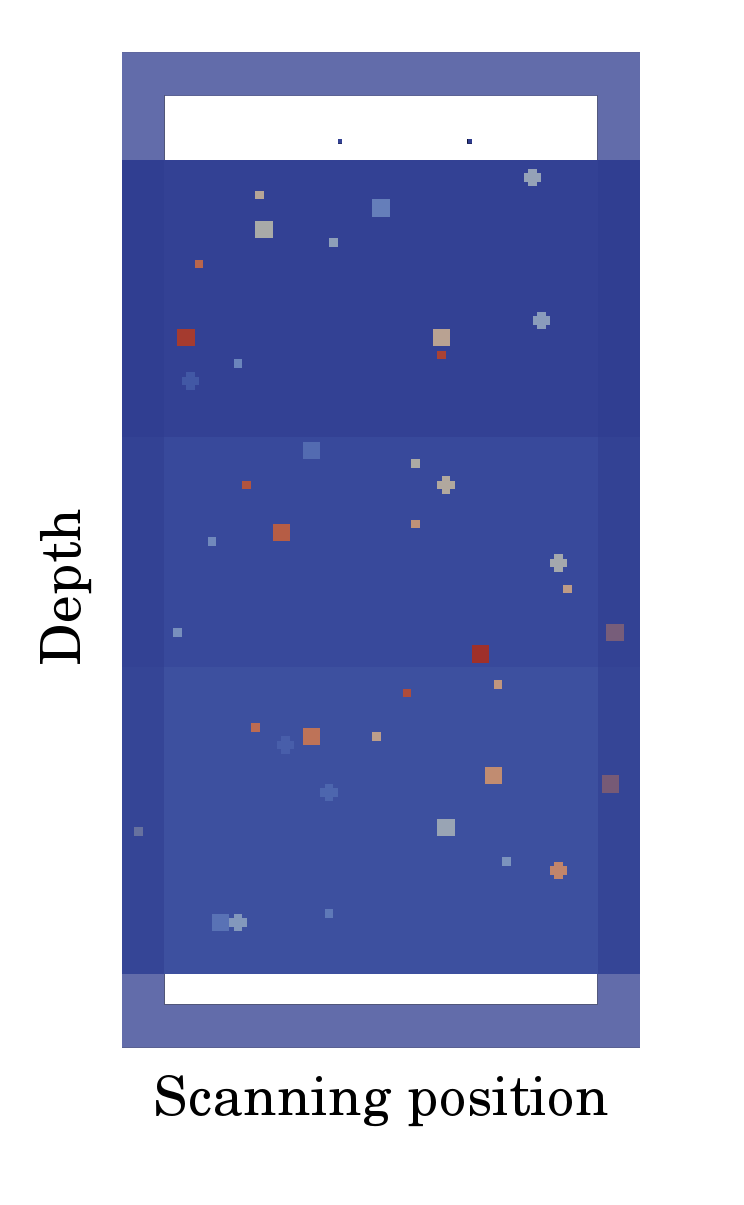}
    \caption{Scene 1}
    \end{subfigure}
    \begin{subfigure}{0.3\textwidth}
    \captionsetup{justification=centering}
    \includegraphics[width=\textwidth]{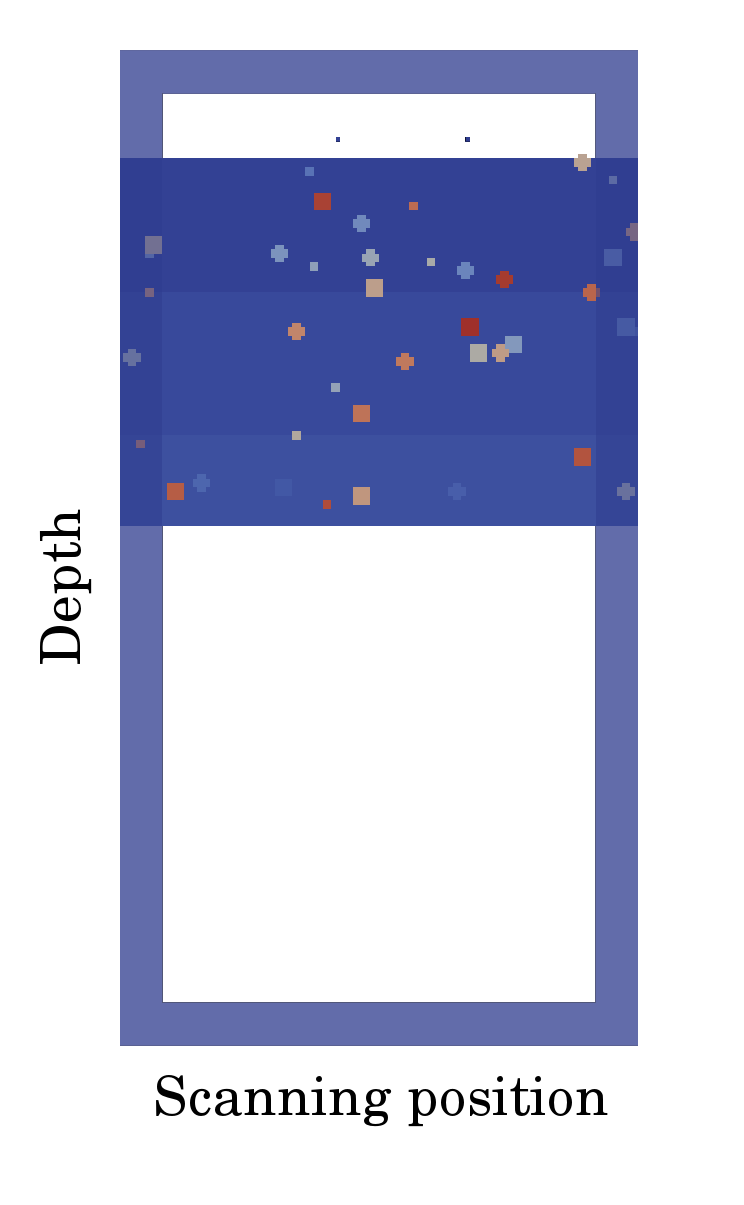}
    \caption{Scene 2}
    \end{subfigure}
    \begin{subfigure}{0.3\textwidth}
    \captionsetup{justification=centering}
    \includegraphics[width=\textwidth]{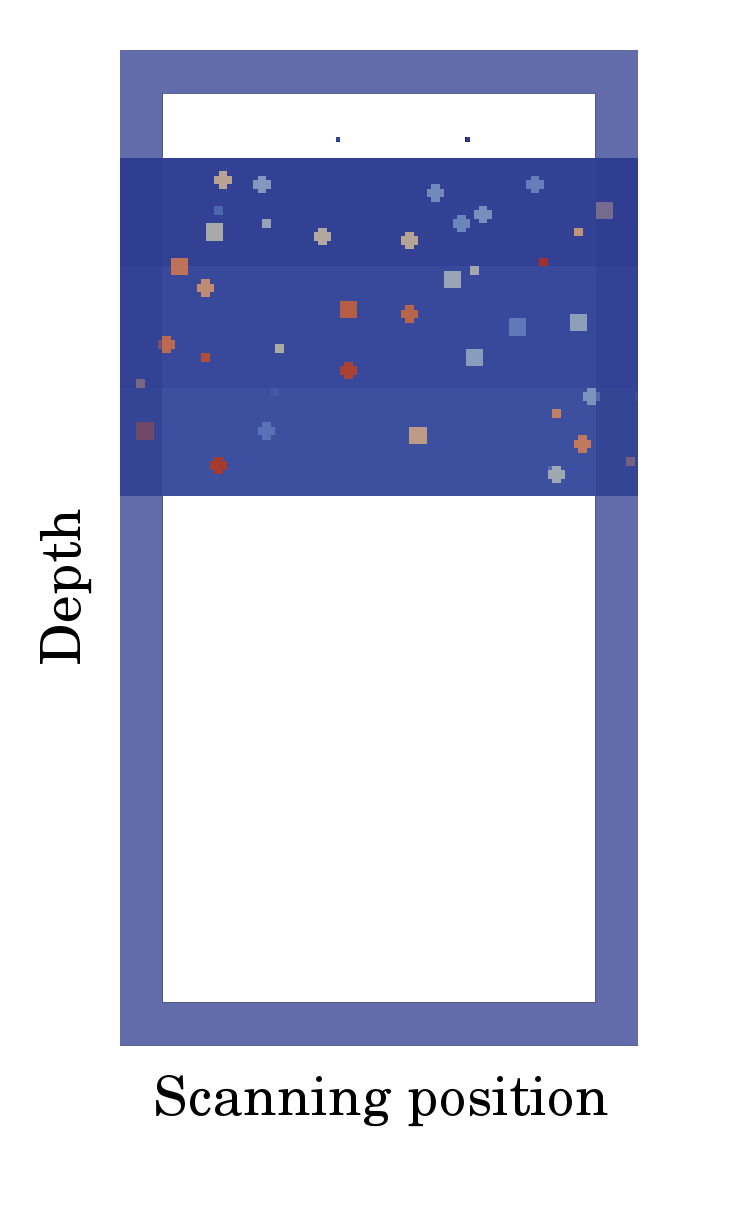}
    \caption{Scene 3}
    \end{subfigure}
    
    \caption{Examples of the simulated B-Scans (a-c) and visualizations of the corresponding setups (d-f)}
    \label{fig:sim_scene}
\end{figure}

Examples of three simulation scenes are shown in Fig. \ref{fig:sim_scene}. They illustrate three-layered structures of different thicknesses and materials. The detailed descriptions of the scenes' parameters are given in Table \ref{tab:three_samples_params}. From the tables and Fig. \ref{fig:sim_scene}, one can see that the relation between the position of the amplitude peaks (a ripple on a radargram) and the actual medium change is not linear. For example, scenes 2 and 3 have very similar layer thicknesses, but due to differences in permittivities, the ripples appear on B-scans 2 and 3 at very different times: two ripples at around 3ns in B-scan 2, one ripple at 3.2ns and another one at 6.2ns in B-scan 3.

\begin{table}[]
    \centering
    \begin{tabular}{|c|c|c|c|}
        \hline
         Thicknesses & Layer 1 & Layer 2 & Layer 3  \\
         \hline
         Scene 1 & 0.128 & 0.105 & 0.143 \\
         \hline
         Scene 2 & 0.063 & 0.065 & 0.042 \\
         \hline
         Scene 3 & 0.05 & 0.05 & 0.05 \\
         \hline
         \addlinespace[2ex]
         \hline
         Permittivities & Layer 1 & Layer 2 & Layer 3  \\
         \hline
         Scene 1 & 4.283 & 4.3725 & 5.516 \\
         \hline
         Scene 2 & 5.423 & 5.226 & 2.991 \\
         \hline
         Scene 3 & 6.278 & 7.584 & 5.841 \\
         \hline
    \end{tabular}
    \captionsetup{justification=centering}
    \caption{Parameters of the scenes' setups in fig. \ref{fig:sim_scene}.}
    \label{tab:three_samples_params}
\end{table}

\section{Results}

The previously described simulated data was used to train the convolutional neural network from the section \ref{cnn}. We split the dataset into training and testing sets in a ratio of 80 to 20. The CNN was trained for 1500 epochs. The network parameters were optimized with the ADAM optimizer with a step length of 0.001.

\begin{figure}[H]
    \centering
    \begin{subfigure}{0.38\textwidth}
    \includegraphics[width=\textwidth]{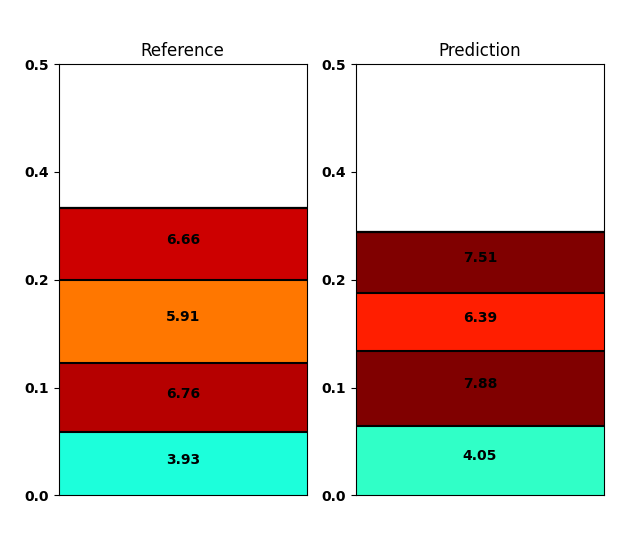}
    \captionsetup{justification=centering}
    \caption{Case 1}
    \end{subfigure}
    \begin{subfigure}{0.38\textwidth}
    \includegraphics[width=\textwidth]{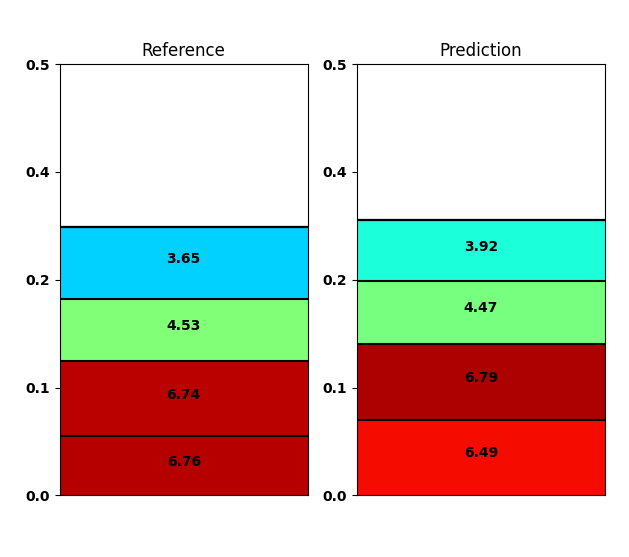}
    \captionsetup{justification=centering}
    \caption{Case 2}
    \end{subfigure}
    \begin{subfigure}{0.38\textwidth}
    \includegraphics[width=\textwidth]{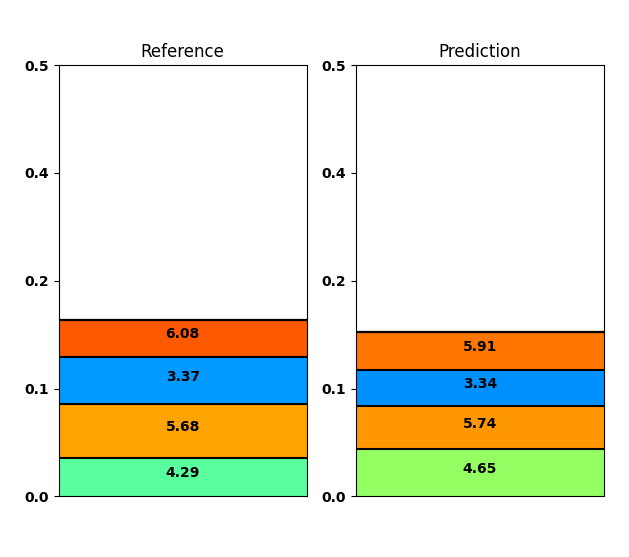}
    \captionsetup{justification=centering}
    \caption{Case 3}
    \end{subfigure}
    \begin{subfigure}{0.38\textwidth}
    \includegraphics[width=\textwidth]{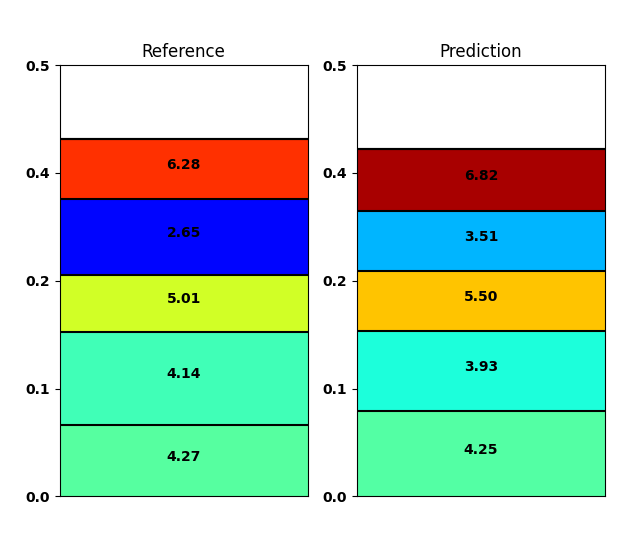}
    \captionsetup{justification=centering}
    \caption{Case 4}
    \end{subfigure}
    \begin{subfigure}{0.38\textwidth}
    \includegraphics[width=\textwidth]{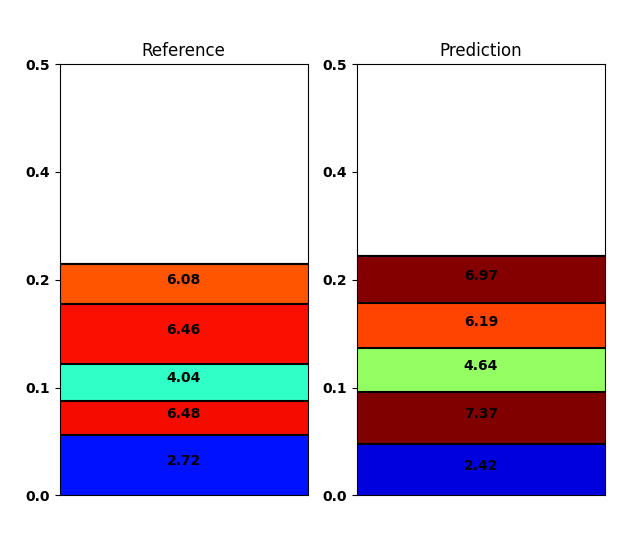}
    \captionsetup{justification=centering}
    \caption{Case 5}
    \end{subfigure}
    \begin{subfigure}{0.38\textwidth}
    \includegraphics[width=\textwidth]{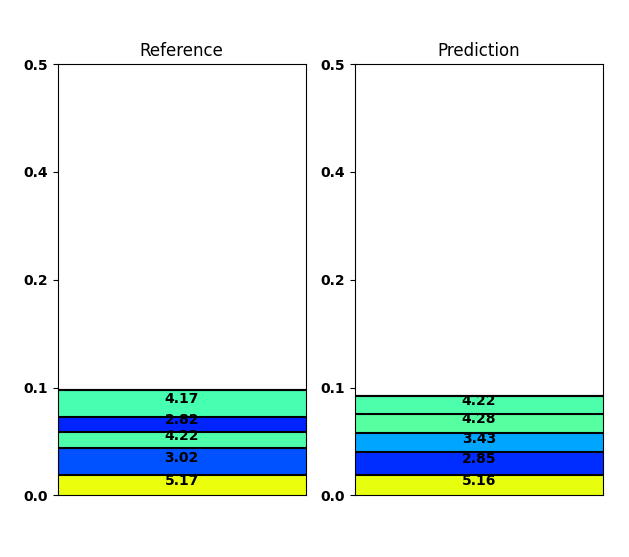}
    \captionsetup{justification=centering}
    \caption{Case 6}
    \end{subfigure}
    
    \vskip\baselineskip
    \begin{subfigure}{0.3\textwidth}
    \includegraphics[width=\textwidth]{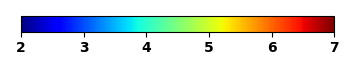}
    \captionsetup{justification=centering}
    \label{fig:colorbar}
    \caption{Colorbar for permittivities}
    \end{subfigure}    
    \captionsetup{justification=centering}
    \caption{Visualization of the prediction results. The blocks represent material layers, ordered by the distance from the scanner bottom to top. The colors correspond to the permittivities of materials with the colorbar depicted in Subfig. (g).}
    \label{fig:examples_visualizations}
\end{figure}

The results of the CNN's performance are visualized in Fig. \ref{fig:examples_visualizations} - where we show six examples from the test dataset (left) together with the output of our trained model (right). We omitted one- or two-layered structures in order to showcase the more complex cases. We observe that low permittivity layers negatively affect the quality of prediction in further layers. Other factors are the thicknesses of layers and their count. 

Statistics over the whole dataset are presented in the Table \ref{tab:sim_dataset_total_stats}, where the average error values in thickness and permittivity are presented. The errors amounted to 9.3 mm in thickness for the training set and 10.2 mm for the test set. Additionally, the error values are presented in percentage against average thicknesses and permittivities of the whole dataset. Table \ref{tab:per_layer_sim_dataset} shows the error values for each layer. It can be observed that the prediction gets worse at deeper layers, which is to be expected, considering the effect of the previous layers. 

\begin{table}[H]
\centering
\begin{tabular}{| c | c | c |} 
 \hline
 \makecell{\textbf{Total mean} \\ \textbf{thickness error} } & average value & \makecell{average value \\ in percentage} \\
 \hline
 & & \\[-1ex]
 \makecell{training} & 9.3 mm & 10.1\% \\[1ex]
 \hline
 & & \\[-1ex]
 \makecell{test} & 10.2 mm & 3.3\% \\[1ex]
 \hline
 \addlinespace[2ex]
 \hline
 \makecell{\textbf{Total mean} \\ \textbf{permittivity error} } & average value & \makecell{average value \\ in percentage} \\
 \hline
 & & \\[-1ex]
 training & 0.71 & 16.4\% \\[1ex]
 \hline
 & & \\[-1ex]
 test & 0.81 & 14.9\% \\[1ex]
 \hline
 
\end{tabular}
\caption{Statistics on the simulated dataset.}
\label{tab:sim_dataset_total_stats}
\end{table}

\begin{table}
\centering
\begin{tabular}{| c | c | c | c | c | c | c |} 
 \hline
  & Layer 1 & Layer 2 & Layer 3 & Layer 4 & Layer 5 & Layer 6 \\
 \hline
 \makecell{mean thickness error \\ (training)} &  15.2\% & 17.9\% & 19.2\% & 21\% & 23.3\% & 25.8\%  \\
 \hline
 \makecell{mean thickness error \\ (test)} &  16.9\% & 19.9\% & 21.5\% & 23.4\% & 26\% & 28.8\%  \\
 \hline
 \makecell{mean permittivity error \\ (training)} &  14.8\% & 14.7\% & 14.6\% & 14.6\% & 14.6\% & 14.7\%  \\
 \hline
 \makecell{mean permittivity error \\ (test)} &  17.2\% & 17.1\% & 17.0\% & 16.9\% & 17.0\% & 17.1\%  \\
 \hline
\end{tabular}
\caption{Thickness and permittivity errors on the simulated dataset for each layer.}
\label{tab:per_layer_sim_dataset}
\end{table}

\subsection{Evaluation on actual buildings}


We collected a set of radargrams from 9 buildings. Each building had walls that could be clustered into groups of a certain build. The builds constitute a composition of layers of a certain thickness and materials. Some builds can have materials that are non-homogenous in their structure, e.g. reinforced concrete or patterned woodwork. Capturing such structures is not within the scope of this project, hence those builds were discarded.

The scans were taken with the pulseEKKO radar equipment with a central frequency of 1 GHz. The preprocessing steps included time-zero calibration and filtering off the frequencies under 500 MHz.

The buildings' walls have various widths, which subsequently results in the non-uniform width of B-scans in the dataset. Therefore, we cut the scans into segments of a fixed length of 40 traces in the width dimension. The cuts are done at random locations within the B-scan, because the scans of the real buildings can contain various artifacts like cables, voids and pipes. By providing several cuts per scan, we let the CNN learn on several samples that represent the same material configuration, thus making it more robust to noise.

As a way of testing the generalization abilities of the trained CNN, we also tried transfer learning. Namely, we used the model that was pre-trained on the simulated dataset and let it train on the collected dataset. Both were trained for 500 epochs. The results are shown in Tables \ref{tab:total_stats_collected} and \ref{tab:transfer_learning_per_layer}. It can be seen that the pre-trained model does not demonstrate better performance. Results on the training part of the collected dataset are slightly better for the untrained model and significantly better on the test dataset for the untrained model too. The reason for the unsuccessful application of transfer learning lies in the deviation between the simulated and the collected data. The task of generating realistic data, that is similar to the output of the target device, is difficult due to the problem of reproduction of the device model. We used the same antenna frequency in the simulation as the device we utilized, but other factors are as important, being it antenna geometry, radiation pattern, etc. These factors made the simulated B-scans considerably different from the ones we collected.

\begin{table}[H]
\centering
\begin{tabular}{| c | c | c |} 
 \hline
 \makecell{\textbf{Total mean} \\ \textbf{thickness error} } & Pre-trained model & Untrained model \\
 \hline
 & & \\[-1ex]
 training & 13.6 mm (11.6\%) & 12.9 mm (10.1\%) \\[1ex]
 \hline
 & & \\[-1ex]
 test & 17.8 mm (7.7\%) & 15.3 mm (2.0\%) \\[1ex]
 \hline
 \addlinespace[2ex]
 \hline
  \makecell{\textbf{Total mean} \\ \textbf{permittivity} } & Pre-trained model & Untrained model \\
 \hline
 & & \\[-1ex]
 training & 0.36 (18.2\%) & 0.17 (14.8\%) \\[1ex]
 \hline
 & & \\[-1ex]
 test & 0.5 (11.8\%) & 0.29 (5.7\%) \\[1ex]
 \hline
 
\end{tabular}
\caption{Results of the transfer learning. The model trained on the collected dataset only shows better performance than the model pre-trained on the simulated dataset.}
\label{tab:total_stats_collected}
\end{table}

The covered materials are listed in Appendix \ref{tab:materials_list}. Some materials that are similar in structure and their dielectric properties were grouped together. The relative permittivities were collected from various sources \cite{p2040}, \cite{ellingson}, \cite{eng_toolbox}.

\begin{table}
\centering
\begin{tabular}{| c | c | c | c | c | c | c |} 
 \hline
 \textbf{Pre-trained model} & Layer 1 & Layer 2 & Layer 3 & Layer 4 & Layer 5 & Layer 6 \\
 \hline
 \makecell{mean thickness error \\ (training)} &  24.4\% & 5.4\% & 18.5\% & 55.1\% & 137.6\% & -  \\
 \hline
 \makecell{mean thickness error \\ (test)} &  38.2\% & 8.5\% & 29.0\% & 86.3\% & 215.4\% & -  \\
 \hline
\end{tabular}
\begin{tabular}{| c | c | c | c | c | c | c |} 
 \hline
 \textbf{Untrained model} & Layer 1 & Layer 2 & Layer 3 & Layer 4 & Layer 5 & Layer 6 \\
 \hline
 \makecell{mean thickness error \\ (training)} &  21.2\% & 4.7\% & 16.1\% & 48.0\% & 119.7\% & -  \\
 \hline
 \makecell{mean thickness error \\ (test)} &  31.0\% & 6.9\% & 23.5\% & 69.9\% & 174.6\% & -  \\
 \hline
\end{tabular}
\caption{Per-layer comparison of the pre-trained model and the untrained (collected data only) model in terms of thickness and permittivity. Lower errors at layer 2 could be related to the fact that the materials at this layer varied less than on the other levels.}
\label{tab:transfer_learning_per_layer}
\end{table}

\begin{table}
\centering
\begin{tabular}{|c|c|c|}
     \hline
     & Pre-trained model & Untrained model \\
     \hline
 & & \\[-1ex]
     accuracy (training) & 90\% & 92\% \\[1ex]
     \hline
 & & \\[-1ex]
     accuracy (test) & 82\% & 90\% \\[1ex]
     \hline
\end{tabular}
\caption{Comparison with the pre-trained model in terms of accuracy of the material classification. Better accuracy is demonstrated by the model trained only on the collected data.}
\label{tab:accuracy}
\end{table}

\section{Conclusion}

In this paper, we investigated the application of deep learning for the non-invasive examination of walls' material composition using GeoRadar. We used a Convolutional Neural Network (CNN) and trained it on a dataset consisting of simulated B-scans. The predictions on the validating dataset have shown promising results, being able to closely identify material in many-layered structures. 
Additionally, the generalization ability of the trained model was tested on the data collected from the real buildings. The experiments have shown little difference between the model trained on the simulated data and the model trained on the real dataset only. However, we could observe high accuracy of the material classification and a good precision of the layers thicknesses prediction. 



\bibliography{bibliography}

\section{Appendix A. Relative permittivities of the considered construction materials}

The relative permittivities that were collected from various sources \cite{p2040}, \cite{ellingson}, \cite{eng_toolbox}.

\begin{table}[H]
\centering
\begin{tabular}{|c | c|} 
 \hline
 material & relative permittivity \\
 \hline
 finery & 5.31 \\
 \hline
 brick & 3.75 \\
 \hline
 bitumen & 2.8 \\
 \hline
 tiles (ceramic) & 21 \\
 \hline
 concrete & 5.31 \\
 \hline
 mineral wool & 1.5 \\
 \hline
 plasterboard & 2.58 \\
 \hline
 steel & 1 \\
 \hline
 heraklith & 1.1\\
 \hline
 ytong & 1.7 \\
 \hline
 styrofoam & 1.06 \\
 \hline
 mortar & 4.7 \\
 \hline

\end{tabular}
\label{tab:materials_list}
\caption{Relative permittivities of the used materials}
\end{table}

\end{document}